\documentclass[10pt,twocolumn,letterpaper]{article}
\usepackage[pagenumbers]{cvpr}
\usepackage{graphicx}
\usepackage{amsmath}
\usepackage{amssymb}
\usepackage{booktabs}
\usepackage{multirow}
\usepackage{bbm}
\usepackage{adjustbox}

\usepackage{amsmath,amsfonts,bm}

\def\eqref#1{equation~\ref{#1}}

\def\1{\bm{1}}

\DeclareMathAlphabet{\mathsfit}{\encodingdefault}{\sfdefault}{m}{sl}
\SetMathAlphabet{\mathsfit}{bold}{\encodingdefault}{\sfdefault}{bx}{n}

\usepackage[usenames,dvipsnames]{xcolor}
\usepackage{xcolor}
\usepackage{pifont}
\usepackage{algorithm}
\usepackage{listings}
\usepackage[accsupp]{axessibility} 
\newcommand{\cmark}{\ding{51}}
\newcommand{\xmark}{\ding{55}}

\usepackage[pagebackref,breaklinks,colorlinks]{hyperref}

\usepackage[capitalize]{cleveref}
\crefname{section}{Sec.}{Secs.}
\Crefname{section}{Section}{Sections}
\Crefname{table}{Table}{Tables}
\crefname{table}{Tab.}{Tabs.}

\def\cvprPaperID{10916} \def\confName{CVPR}
\def\confYear{2022}

\usepackage{overpic}
\usepackage{enumitem} 
\usepackage{overpic} 
\usepackage{color}

\definecolor{turquoise}{cmyk}{0.65,0,0.1,0.3}
\definecolor{purple}{rgb}{0.65,0,0.65}
\definecolor{dark_green}{rgb}{0, 0.5, 0}
\definecolor{orange}{rgb}{0.8, 0.6, 0.2}
\definecolor{red}{rgb}{0.8, 0.2, 0.2}
\definecolor{darkred}{rgb}{0.6, 0.1, 0.05}
\definecolor{blueish}{rgb}{0.0, 0.3, .6}
\definecolor{light_gray}{rgb}{0.7, 0.7, .7}
\definecolor{pink}{rgb}{1, 0, 1}
\definecolor{greyblue}{rgb}{0.25, 0.25, 1}

\usepackage{blindtext}

\renewcommand{\paragraph}[1]{\vspace{1em}\noindent\textbf{#1}.}
\begin{document}

\title{Dual Temperature Helps Contrastive Learning Without Many Negative Samples: \\ Towards Understanding and Simplifying MoCo}

\author{Andrea Tagliasacchi\\
Google Research \& University of Toronto\\
{\tt\small taglia@google.com}

}

\author{
      Chaoning Zhang$^{1}$\thanks{equal contribution. corresponding author: Chaoning Zhang chaoningzhang1990@gmail.com}, Kang Zhang$^{1,*}$, Trung X. Pham$^{1,*}$, Axi Niu$^{2}$, Zhinan Qiao$^{3}$ \\
        Chang D. Yoo$^{1}$, In So Kweon$^{1}$  \\

     $^{1}$ KAIST,
     $^{2}$ Northwestern Polytechnical University,
     $^{3}$ University of North Texas\\

}

\maketitle

\begin{abstract}

Contrastive learning (CL) is widely known to require many negative samples, 65536 in MoCo for instance, for which the performance of a dictionary-free framework is often inferior because the negative sample size (NSS) is limited by its mini-batch size (MBS). To decouple the NSS from the MBS, a dynamic dictionary has been adopted in a large volume of CL frameworks, among which arguably the most popular one is MoCo family. In essence, MoCo adopts a momentum-based queue dictionary, for which we perform a fine-grained analysis of its size and consistency. We point out that InfoNCE loss used in MoCo implicitly attract anchors to their corresponding positive sample with various strength of penalties and identify such inter-anchor hardness-awareness property as a major reason for the necessity of a large dictionary. Our findings motivate us to simplify MoCo v2 via the removal of its dictionary as well as momentum. Based on an InfoNCE with the proposed dual temperature, our simplified frameworks, SimMoCo and SimCo, outperform MoCo v2 by a visible margin. Moreover, our work bridges the gap between CL and non-CL frameworks, contributing to a more unified understanding of these two mainstream frameworks in SSL. Code is available at: \url{https://bit.ly/3LkQbaT}.

\end{abstract}

\section{Introduction}

Self-supervised learning (SSL) has become increasingly popular in various domains, ranging from NLP~\cite{Lan2020ALBERT,radford2019language,devlinetal2019bert,su2020vlbert,nie2020dc} to visual representation~\cite{oord2018representation,he2020momentum,chen2020simple}, in the past few years.
Especially, contrastive learning (CL)  frameworks~\cite{oord2018representation,hjelm2018learning,wu2018unsupervised,zhuang2019local,bachman2019learning,henaff2020data,tian2020contrastive,chen2020simple,he2020momentum,wang2020understanding,wang2020DenseCL} have attracted significant attention due to its intuitive motivation. 
In essence, CL is designed to attract the anchor sample~\cite{Wang_2021_CVPR} close to the positive sample, \ie\ another augmented view of the same image, and simultaneously repulse it from negative samples, \ie\ views from different images. 
With the popular InfoNCE loss~\cite{oord2018representation}, CL is widely reported to require a large amount of negative samples~\cite{he2020momentum}. For example, ~\cite{chen2020simple} shows that increasing the mini-batch size (MBS) to a large value, 4096 for instance, is essential for achieving competitive performance, for which there are multiple challenges, such as GPU memory concern or difficulty to train with a large MBS~\cite{you2017large,chen2020simple}.
Thus, a major line of CL frameworks, such as MoCo~\cite{he2020momentum}, have emerged to decouple the required large negative sample size (NSS) from the MBS with a dynamic dictionary. Despite much effort in the dictionary design~\cite{wu2018unsupervised,he2020momentum}, 
\textit{why contrastive InfoNCE requires a large dictionary (or many negative samples) is not well understood}.

\begin{figure}[t]
  \centering
 \includegraphics[width=1\linewidth]{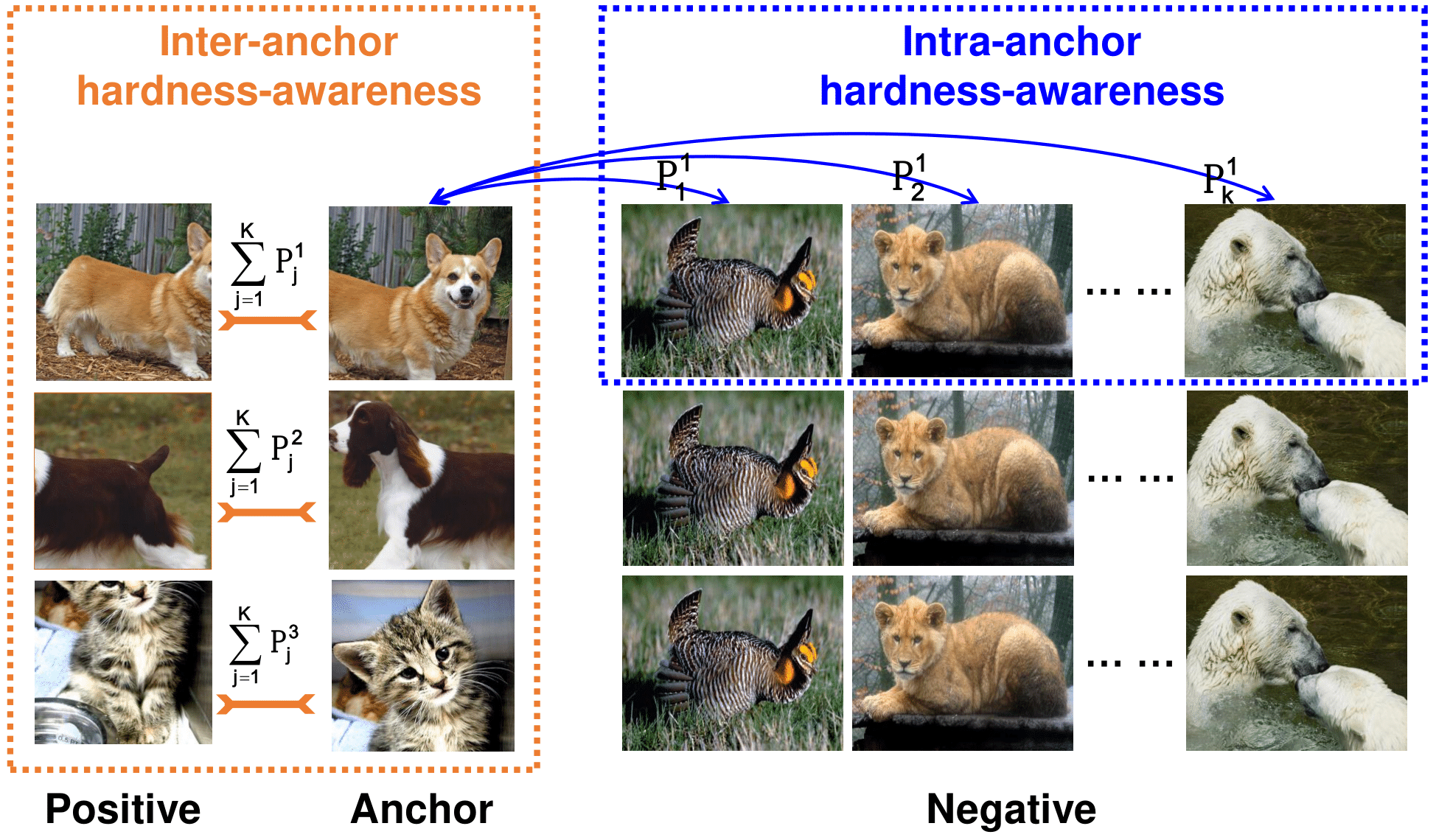}
  \caption{
  \textcolor{blue}{Intra-anchor} and \textcolor{YellowOrange}{Inter-anchor} hardness-aware properties. The former is indicated by different repulsing weights (see Eq~\ref{eq:prob}) for different negative samples based on their hardness, \ie\ $\bm{p}^1_1 \neq \bm{p}^1_2 ... \neq \bm{p}^1_K$ ($K$ denotes NSS), and the latter is indicated by different weights being put on different  anchor samples to attract the corresponding positive sample, \ie\ $\sum_{j=1}^{K} \bm{p}^1_j \neq \sum_{j=1}^{K} \bm{p}^2_j \neq \sum_{j=1}^{K} \bm{p}^3_j$ with three anchor images as a motivation example.
  }
  \label{fig:teaser}
\end{figure}

Our investigation of the above problem centers around an interesting hardness-aware property~\cite{Wang_2021_CVPR} of InfoNCE. A large volume of works~\cite{nozawa2021understanding, Zeng_2020_CVPR, Vasudeva_2021_ICCV, Bai_2021_ICCV, iscen2018mining, ho2020contrastive} have studied strategies of mining hard negative samples, \ie\ those samples that are similar to the anchor sample. We point out that the anchor sample also has this hardness property. Conceptually, an anchor sample is considered hard when it is still far from the positive sample and/or close to negative samples.  InfoNCE loss has been identified to have the hardness-aware property~\cite{Wang_2021_CVPR}, which contributes to dimensional de-correlation~\cite{zhang2022how}, is critical for performance. 

Prior works~\cite{Wang_2021_CVPR,zhang2022how} mainly study the hardness-awareness \textit{within an anchor}, which is therefore termed \textit{intra-anchor} hardness-aware property here. As in Figure~\ref{fig:teaser}, it indicates that the gradient puts different weights (see Eq~\ref{eq:prob}) on various negative samples for repulsing the anchor from them with different strength of penalties, \ie\ $\bm{p}^i_1 \neq \bm{p}^i_2 ... \neq \bm{p}^i_K$ (the fixed superscript $i$, 1 for instance, denotes the same anchor). In contrast, the \textit{inter-anchor} hardness-aware property indicates different weights being on anchors for attracting them to their corresponding positive sample with different penalties, \ie\ $\sum_{j=1}^{K} \bm{p}^1_j \neq \sum_{j=1}^{K} \bm{p}^2_j \neq \sum_{j=1}^{K} \bm{p}^3_j$ with three anchors as an example.

Overall, our contributions are summarized as follows. 

\begin{itemize}

    \item We point out that anchors have hardness property, for which contrastive InfoNCE loss by default attracts them to their corresponding positive samples with various strength of penalties.
    Recognizing this, we disentangle InfoNCE into vector and scalar components that reflect intra-anchor and inter-anchor hardness-aware properties, respectively. Such a decomposed loss facilitates a fine-grained analysis on the MoCo dictionary and we reveal: (i) a small dictionary is sufficient for the vector component which, however, requires high consistency between encoders for representing the negative and positive keys; (ii) the scalar component requires a very large dictionary but is less sensitive to such consistency.

    \item We identify that the increase of dictionary size and temperature both help alleviate the inter-anchor hardness-aware sensitivity for improving performance. Our findings help simplify MoCo family via removing their dictionary and momentum. Specifically, we propose dual temperature for realizing independent control of intra-anchor and inter-anchor properties. Without a dictionary, our proposed SimMoCo achieves comparable or superior performance over the baseline MoCo v2. Notably, our dictionary-free and momentum-free SimCo is simple yet effective. 
    
    \item Our investigation helps bridge the gap between CL and non-CL frameworks, contributing to a unified perspective on these two major SSL frameworks.

\end{itemize}

\section{Related Work} \label{sec:related}
Recently, multiple works~\cite{chen2021exploring,grill2020bootstrap,ermolov2021whitening,bardes2021vicreg,zbontar2021barlow} have attempted SSL without using negative samples, demonstrative performance comparable to the CL frameworks. However, they are often dependent on additional predictor~\cite{grill2020bootstrap} or stop gradient~\cite{chen2021exploring} or explicit de-centering and de-correlation~\cite{ermolov2021whitening,bardes2021vicreg,zbontar2021barlow,zhang2022how}. CL remains as a mainstream framework for SSL and has also been extensively studied in other filed applications \cite{Eun_2020_CVPR, Zhuang_2020_CVPR, Wu_2021_CVPR, Pan_2021_CVPR, Wang_2021_CVPR, Aberdam_2021_CVPR, Yao_2021_CVPR, Yu_2021_ICCV, Hu_2021_ICCV, Diba_2021_ICCV, Xie_2021_ICCV}.

\textbf{Contrastive learning.} The core of unsupervised learning is to learn augmentation-invariant representation, for which CL is at the core of its development~\cite{Schroff2015FaceNetAU,wang2015unsupervised,sohn2016improved,misra2016shuffle,Federici2020Learning}. Inspired by this success, CL has been extensively studied for SSL pretext training~\cite{wu2018unsupervised,oord2018representation,bachman2019learning,henaff2020data,hjelm2018learning,tian2019contrastive,zhuang2019local,chen2020simple}. Early works have attempted margin-based contrastive losses~\cite{hadsell2006dimensionality,wang2015unsupervised,hermans2017defense} and \cite{wu2018unsupervised,oord2018representation} propose a NCE-like loss which has become the de facto standard loss in CL.

More recently, demonstrating superior performance over supervised pre-training counterparts, MoCo~\cite{he2020momentum} has attracted significant attention. 
MoCo v2~\cite{chen2020mocov2} incorporates stronger augmentation and additional MLP projector head from~\cite{chen2020simple}, which shows significant performance improvement over MoCo v1. Moreover, ~\cite{chen2021empirical} has demonstrated that MoCo family can also exploit ViT structures~\cite{dosovitskiy2021an} based on which they find that prediction head from the non-CL frameworks~\cite{chen2021exploring,grill2020bootstrap} brings additional performance boost. In essence, what is unique to MoCo family is their dictionary, where the keys are also found to benefit from increased diversity through negative interpolation~\cite{zhu2021improving}. The understanding of this core component, \ie\ momentum-based queue dictionary, in MoCo is limited and our work fills the gap to perform a fine-grained analysis.

A key property of CL is that it involves negative samples, and a major line works~\cite{nozawa2021understanding, Zeng_2020_CVPR, Vasudeva_2021_ICCV, Wang_2021_ICCV, Bai_2021_ICCV, iscen2018mining, chuang2020debiased, ho2020contrastive,wu2020conditional,NEURIPS2020_f7cade80,robinson2021contrastive} have shown that mining hard negative samples can be beneficial for performance. Moreover, ~\cite{Wang_2021_CVPR} has identified that InfoNCE has a hardness-aware property which is critical for competitive performance.~\cite{zhang2022how} has shown that this can be attributed to the effect of dimensional de-correlation. In contrast to them that mainly focused on intra-anchor hardness-awareness, our work studies the inter-anchor hardness-aware property and identifies it as a major reason to explain why MoCo family requires a large dictionary.

\textbf{Temperature in CL.}

Temperature plays a key role for the success of CL due to its \textit{hardness-aware}~\cite{Wang_2021_CVPR} or de-correlation~\cite{zhang2022how}. They analyze the influence of the temperature in the vector component, while ours is the first to decompose the influence of temperature into two components. The dual temperature has been previously studied in a non-CL framework termed DINO~\cite{caron2021emerging} from the perspective of knowledge distillation. Specifically, the teacher adopts a lower temperature than that of the student for help distilling knowledge. By contrast, our work adopts dual temperature in a contrastive InfoNCE for realizing independent control of two hardness-aware sensitiveness. 
Recently,~\cite{zhang2021temperature} has also exploited input-dependent learnable temperature in SSL for estimating uncertainty in out-of-distribution detection. It might be interesting to apply our dual temperature concept to~\cite{zhang2021temperature} for identifying which one (or both) is beneficial for such uncertainty estimation.

\section{Background} \label{sec:background}

\textbf{A large dictionary is desirable.} 
Driven by various motivations, multiple works~\cite{wu2018unsupervised,oord2018representation,bachman2019learning,henaff2020data,hjelm2018learning,tian2019contrastive,zhuang2019local} have designed dynamic dictionaries and exploited the stored keys as negative samples. This dictionary is desirable to be large, for which~\cite{wu2018unsupervised} proposes to save the representations of all training samples in a memory bank. To increase the consistency among the stored representations, MoCo~\cite{he2020momentum} proposes a FIFO queue dictionary based on the momentum encoder. The influence of such consistency on MoCo is demonstrated in~\cite{zhu2021improving} by analyzing the effect of momentum coefficient. Without a dictionary, the negative sample size would be limited by the MBS. The main merit of a dictionary 
is to decouple the NSS from the MBS, which allows access to a large number of negative samples without increasing the MBS. Despite many attempts at exploring various dictionaries, less attention has been paid to understanding why CL requires a large dictionary.

\textbf{Contrastive loss.} NCE-like loss~\cite{gutmann2010noise} has been independently introduced with various motivations in multiple popular works~\cite{sohn2016improved,wu2018unsupervised,oord2018representation} and it has emerged as the de-facto standard loss for CL. Following~\cite{oord2018representation,he2020momentum,zhang2022how}, we term it InfoNCE for consistency. Given an encoder $f$, a random input sample $\bm{x}$ is encoded as a query (or anchor) $q$ or key $k$, which are often $l_2$ normalized to avoid scale ambiguity. We consider a set of encoded queries (anchors) $\{q_1, q_2, ...\}$ and encoded keys $\{k_1, k_2, ...\}$. With similarity measure by dot product, InfoNCE is formulated as: 
\begin{equation}
    \mathcal{L}_{q_{i}} = -\log \frac{\exp(q_i{\cdot}k_+/\tau)}{\exp(q_i{\cdot}k_+/\tau) + \sum_{j=1}^K \exp(q_i{\cdot}k_j/\tau)}
    \label{eq:infonce}
\end{equation}
where 
$k_+$ denotes the positive key to anchor $q_i$ and $\tau$ denotes the temperature. 
This loss has low value when $q_i$ is similar to its positive key and dissimilar to negative keys.

\textbf{Hardness-aware property.} To guide an anchor close to its positive key and far from negative keys, a simple loss as
\begin{equation}
    \mathcal{L}_{simple} = -q_i{\cdot}k_+ + \frac{1}{K}\sum_{j=1}^K q_i{\cdot}k_j,
    \label{eq:simple}
\end{equation}
has been designed in~\cite{Wang_2021_CVPR}. The gradient on $q_i$ is derived as
\begin{equation}
    \frac{ \partial \mathcal{L}_{simple} }{\partial q_i}  = -(k_+ - \frac{1}{K}\sum_{j=1}^K k_j),
    \label{eq:simple_grad}
\end{equation}
which shows that the penalty weight on each negative key is the same.~\cite{Wang_2021_CVPR} has identified that InfoNCE outperforms the above simple loss due to its hardness-aware property via putting more penalty weight on those hard keys. This is reflected in the derived gradient of Eq~\ref{eq:infonce} on $q$ as
\begin{equation}
    \frac{\partial\mathcal{L}_{q_i}}{\partial q_i} = - \left( (\sum_{j=1}^{K} \bm{p}^i_j) k_+ - {\sum_{j=1}^{K} } \bm{p}^i_j k_j \right),
    \label{eq:infonce_grad}
\end{equation}
where a constant component $\frac{1}{\tau}$ is omitted for simplicity because it can be perceived as part of the learning rate. $\bm{p}^i_j$ conceptually indicates the probability of $q_i$ being recognized as $k_j$, which is detailed as
\begin{equation}
    \bm{p}^i_j = \frac{\exp({q_i {\cdot} {k}_j / \tau})}{\exp(q_i{\cdot}k_+/\tau) +  \sum_{r=1}^{K} \exp({q_i \cdot k_r/\tau})}.
    \label{eq:prob}
\end{equation}
Note that for a fixed query $\bm{q}_i$, $\bm{p}^i_j$ ($j \in [1,K]$) are those weights in the intra-anchor hardness-awareness of Fig~\ref{fig:teaser}. Proportional to $\exp({q_i {\cdot}  {k}_j / \tau})$, $\bm{p}^i_j$ indicates more penalty weight being put on hard negative samples~\cite{Wang_2021_CVPR}.

\section{Towards Understanding MoCo}
The core of the seminal MoCo centers around its momentum encoder (MoEn)-based dictionary~\cite{he2020momentum}. Our work revisits a prior hypothesis and performs a fine-grained analysis on its dictionary for a new understanding of MoCo.

\textbf{Prior hypothesis.} It is hypothesized that the dictionary needs to be \textit{large} and \textit{consistent}~\cite{he2020momentum}. Regarding the size, it is assumed in~\cite{he2020momentum} that ``Intuitively, a larger dictionary may better sample the underlying continuous, high-dimensional visual space". Straightforwardly,~\cite{he2020momentum} attributes the necessity of a large dictionary size to the requirement of sampling. Indeed, as they claim, in general, it is \textit{intuitive} that more samples are necessary for better modelling a more high-dimensional space. However, whether this is indeed the major reason for the requirement of a large dictionary remains unclear. On the other hand, regarding the consistency, it is argued in~\cite{he2020momentum} that ``the keys in the dictionary should be represented by the same or similar encoder so that their comparisons to the query are consistent". The stored keys in the dictionary from the past iterations are used as negative keys, thus the necessity of MoEn was mainly attributed to such negative-negative (NN) consistency. Here, we attempt to examine the above claims with a focus on InfoNCE's hardness-aware property.

\subsection{Inter-Anchor Hardness-Aware Property}
It has been noted in~\cite{Wang_2021_CVPR} that in the gradient on the anchor, the weight on the positive key is equal to the sum of weights on all negative keys, \ie\ $\sum_{j=1}^{K}\bm{p}^i_j$ (see Eq~\ref{eq:infonce_grad}). Prior work is mainly interested in the unevenness of those $\bm{p}^i_j$ within an anchor. Our work pays attention to the value of this sum and it motivates us to decompose Eq~\ref{eq:infonce_grad} as 
\begin{equation}
    \frac{\partial\mathcal{L}_{q_i}}{\partial q_i} = - \underbrace{ {\sum_{j=1}^{K}\bm{p}^i_j} }_{ \substack{ \text{Inter-anchor} \\ \text{hardness-awareness} } }  (k_+ - \sum_{j=1}^{K} \underbrace{ \hat{\bm{p}}^i_j }_{ \substack{  \text{Intra-anchor} \\ \text{ hardness-awareness}  }} k_j) ,
    \label{eq:infonce_grad_decomp}
\end{equation}
where $\hat{\bm{p}}^i_j = \bm{p}^i_j/\sum_{j=1}^{K}\bm{p}^i_j$. With such decomposition, we note that the weight on positive key is still equal to the sum of $\hat{\bm{p}}^i_j$ ( $1=\sum_{j=1}^{K}\hat{\bm{p}}^i_j$). Clearly, $\sum_{j=1}^{K}\bm{p}^i_j$ is an anchor-wise weight for indicating the hardness of anchor $q_i$ (see Fig.~\ref{fig:teaser}).

\textbf{Loss transformation.} We reformulate Eq~\ref{eq:infonce} as: 
\begin{equation}
    \mathcal{L}^{new}_{q_i} =  \text{sg}[\frac{\partial\mathcal{L}_{q_i}}{\partial q_i}] \cdot  q_i = \underbrace{\text{sg}[ {\sum_{j=1}^{K}\bm{p}^i_j]} }_{\text{Scalar component}} q_i \cdot \underbrace{\text{sg}[(k_+ - {\sum_{j=1}^{K}} \hat{\bm{p}}^i_j k_j)]}_{\text{Vector component}} ,
    \label{eq:infonce_nss_equiva}
\end{equation}
where $\text{sg}[\cdot]$ indicates the \textit{stop gradient}. Intra-anchor and inter-anchor hardness-awareness are reflected in the vector and scalar components, respectively. This loss transformation enables the adoption of independent dictionaries for the two components. The above loss is mathematically equivalent to that in Eq~\ref{eq:infonce} for optimizing $q$ because they share the same gradient on $q$ (omitted constant $\frac{1}{\tau}$ is considered in practical implementation). Note that $k$ in Eq~\ref{eq:infonce} has no gradients because they are from the dictionary or the output of the momentum encoder~\cite{he2020momentum}.

The above gradient decomposition and loss transformation facilitate the analysis of the dictionary. Our following analysis is based on MoCo v2, however, for concept and notation simplicity, it is still referred to as MoCo.

\subsection{A Fine-Grained Analysis on MoCo Dictionary} \label{sec:finegrained}

\begin{figure}[!htbp]
  \centering
 \includegraphics[width=0.8\linewidth]{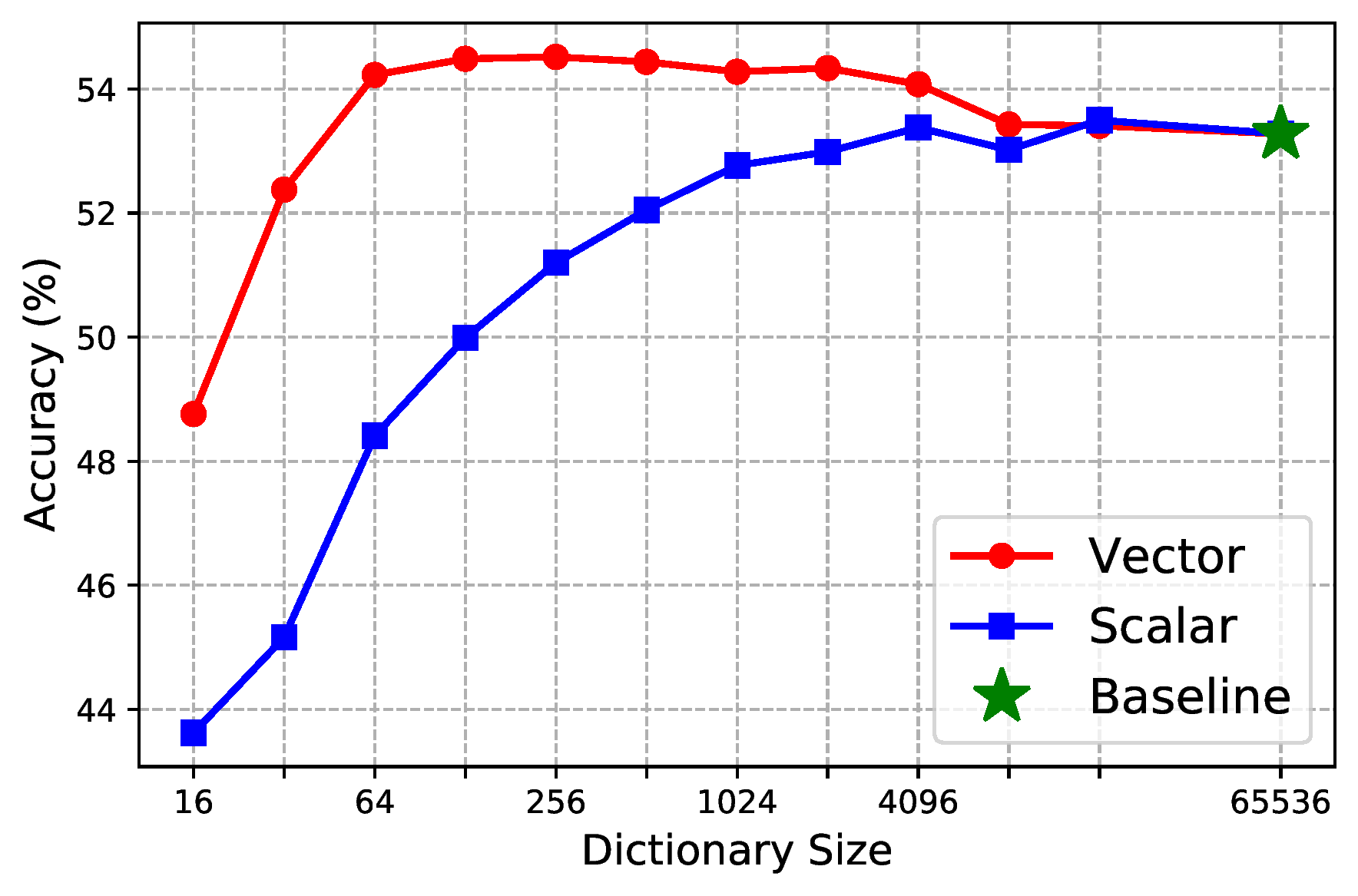}
  \caption{Influence of the size of a independent dictionary. The star indicates the baseline setting the size of both $\bm{D}_{vector}$ and $\bm{D}_{scalar}$ to 65536. Fixing the $\bm{D}_{scalar}$ to 65536, the red line shows the influence of $\bm{D}_{vector}$ size. Fixing the $\bm{D}_{vector}$ to 65536, the blue line shows the influence of $\bm{D}_{scalar}$ size. The experiments are performed on CIFAR100 with MoCo v2 for 200 epochs. Detailed setup is included in the \textcolor{black}{supplementary}.}

  \label{fig:interintra_moco2}
\end{figure}

\textbf{Size.} By default, MoCo adopts a very large dictionary size, 65536 for instance, and we treat it as the baseline of our investigation. Since CL requires a large dictionary size, decreasing the size is confirmed to decrease the performance. 
The dictionary size has an influence on both vector and scalar components. To disentangle such influence, based on the loss in Eq~\ref{eq:infonce_nss_equiva}, we adopt two independent dictionaries, $\bm{D}_{scalar}$ and $\bm{D}_{vector}$, for the scalar and scalar components, respectively. We investigate two scenarios: (a) adopting various $\bm{D}_{scalar}$ sizes with the $\bm{D}_{vector}$ size fixed to 65536; and (b) adopting various $\bm{D}_{vector}$ sizes with the $\bm{D}_{scalar}$ size fixed to 65536. 

From the results in Figure~\ref{fig:interintra_moco2}, there are two major observations. First, the scalar component is highly sensitive to its dictionary size and the performance is much worse when $\bm{D}_{scalar}$ is small. Second, a dictionary size as small as 64 in the vector component is already sufficient for competitive performance. Interestingly, for the vector component, the performance is optimal when the dictionary size is around 256, \ie\ only the keys stored in the last iteration is used since the MBS is set to 256 in this setup.

\begin{table}[!htbp]
    \centering
    \begin{tabular}{ccccc}
    \toprule
    Sampling strategies & Earliest & Random & Newest \\
    \midrule
    Top-1 Accuracy(\%) &1  & 44.03 &  53.43 \\
                                                \bottomrule
    \end{tabular}
    \caption{Comparison of various sampling strategies on CIFAR100}         
    \label{tab:comp_quality}
\end{table}

\textbf{Consistency.} 
~\cite{he2020momentum} mainly attributes the quality of stored keys to NN consistency. By contrast, we conjecture that it might be more important for the positive and negative keys to be represented by the same or similar encoders. Since the keys from the current MoEn are used as the positive one, it is straightforward that the stored order might be an important factor with positive-negative (PN) consistency considered. 

To verify our conjecture, with a full $\bm{D}_{scalar}$ used, we sample $K_{vector}$ keys from the $\bm{D}_{vector}$. The investigated sampling strategies are as follows: (a) sampling the earliest $K_{vector}$ keys; (c) randomly sampling $K_{vector}$ keys; (c) sampling the most recent $K_{vector}$ keys. Recall that adopting 256 keys for the vector component actually outperforms that with a very large dictionary, and we set $K_{vector}$ to 256. Sampling the earliest $K_{vector}$ keys guarantees high NN consistency, however, the results in Table~\ref{tab:comp_quality} show that it leads to non-convergence. With such a sampling strategy, we confirm that increasing $K_{vector}$ to a much larger value, 4096 for instance, does not alleviate such collapse. On the other hand, random sampling has very low NN consistency but high PN consistency leads to a reasonable performance but under-performs that with the most recent keys. Overall, the results suggest that (a) PN consistency better indicates the quality of the stored keys; (b) the vector component is sensitive to such consistency. This also helps explain the interesting phenomenon in Figure~\ref{fig:interintra_moco2} that adopting keys only from the last iteration outperforms that with a very large dictionary. For the vector component, \textit{a large dictionary is not always optimal from the perspective of PN consistency since it contains many old keys}. By contrast, a larger dictionary in the scalar component consistently improves the performance (see Figure~\ref{fig:interintra_moco2}), suggesting the scalar component is less sensitive to the quality of the keys. A more detailed discussion on this is in the \textcolor{black}{supplementary}.

\textbf{Relation to prior hypothesis.} (i) Regarding the size, it might be tempting to believe that the vector component, which matches the high-dimensional representation space, is more sensitive to the dictionary size. Our results show that a relatively small dictionary size is sufficient for the vector component. 
(ii) Regarding the consistency, as discussed above, our work shows that their suggested NN consistency~\cite{he2020momentum} is less informative than PN consistency for indicating the quality of the keys. The importance of PN consistency also helps justify their FIFO queue strategy~\cite{he2020momentum}.

\section{Towards Simplifying MoCo}
The dictionary requires additional memory to store negative keys and the keys need to be encoded by a \textit{momentum encoder} to increase their consistency (or quality) in MoCo~\cite{he2020momentum}. Such dictionary and momentum increase the framework complexity, which motivates us to check the possibility to remove them without performance drop. Somewhat surprisingly, our proposed simplified frameworks actually achieve superior performance over the baseline MoCo v2. The simplifying procedure, as well as the underlying rationale are detailed in the following.

\subsection{Dictionary Removal}
For investigating the role of temperature in controlling the strength penalties on negative sample, ~\cite{Wang_2021_CVPR} defines $r_i(s_{i,j}) = \bm{p}^i_j/\sum_{j=1}^{K}\bm{p}^i_j$, \ie\ $\hat{\bm{p}}^i_j$, as the relative penalty on negative key $k_j$ for the anchcor $q_i$ and analyzes its entropy. Inspired by it, we define $r^i_+$ as the relative penalty weight on anchor $q_i$ to attract their corresponding positive sample:
\begin{equation}
    r^i_+ = \sum_{j=1}^{K}\bm{p}^i_j/\sum_{i=1}^{N} \sum_{j=1}^{K}\bm{p}^i_j.
    \label{eq:inter_penalty}
\end{equation}

\begin{figure}[!htbp]
  \centering
 \includegraphics[width=0.8\linewidth]{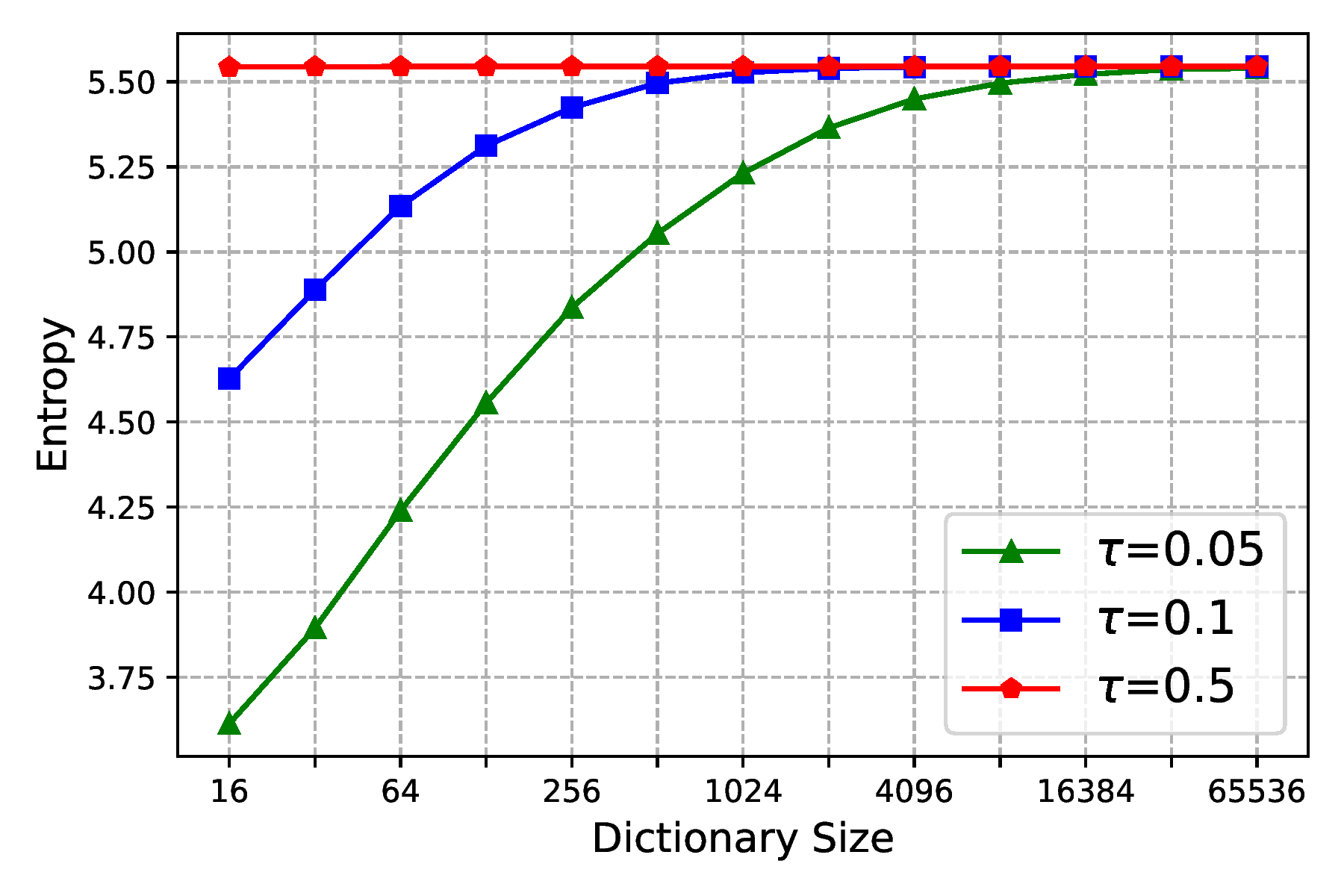}
  \caption{Entropy of $r_+$ under various dictionary sizes and temperature $\tau$ values.}
  \label{fig:entropy}
\end{figure}

From the results in Figure~\ref{fig:entropy}, we observe that the entropy of $r_+$ consistently decreases as the dictionary size decreases. On the other hand, its entropy also decreases when the temperature is set lower. A lower entropy indicates the relative penalty weight on each anchor is less equal, \ie\ more relative weight on the hard anchors. This mirrors the finding in~\cite{Wang_2021_CVPR} that a low temperature decreases the entropy of the $r_i(s_{i,j})$ causing more penalty on the hard negative samples. From this observation, we conjecture that the performance with a small dictionary in the scalar component might benefit from a larger temperature similar to the performance boost caused by a larger dictionary size. Temperature in the vector and scalar components are denoted by $\tau_{\alpha}$ and $\tau_{\beta}$, respectively. To exclude the influence of temperature change in the vector component, we keep $\tau_{\alpha} = 0.1$ fixed and only change the value of $\tau_{\beta}$. The dictionary size is set to 256 for both scalar and vector components. The results are shown in thea first row of Table~\ref{tab:temp_abla}.
\begin{table}[!htbp]
    \centering
    \begin{tabular}{cccccc}
    \toprule
    \multirow{2}{*}{Key type} &\multicolumn{5}{c}{Scalar-temperature ($\tau_{\beta}$) } \\
    &0.05&0.1 & 0.5 & 1.0 & 5 \\
    \midrule
                Last & 43.56 & 49.74 & 54.07 & 54.03 & 53.89 \\
    Current & 44.19 & 49.89 & 54.45 & 54.11 & 54.0 \\
    \bottomrule
    \end{tabular}
    \caption{Influence of scalar temperature $\tau_{\beta}$ on CIFAR100. $\tau_{\alpha}=0.1$ in all experiments here.} 
    \label{tab:temp_abla}
\end{table}

\textbf{SimMoCo.} We observe that a low temperature, 0.05 for instance, leads to a significant performance drop (43.56\%), while a sufficiently large temperature is beneficial for superior performance. Note that by default MoCo does not use the current mini-batch keys as the negative samples. Thus the above-discussed results are based on the negative keys saved in the dictionary from the last iteration. Straightforwardly, the keys at the current step from the momentum encoder can be used to replace the above negative keys. The results in the second row of Table~\ref{tab:temp_abla} show that this replacement increases the performance. This (slight) performance boost can be attributed to the improved positive-negative consistency (note that the positive keys are from the momentum encoder at the current iteration). Through this replacement, we show that the dictionary in MoCo can be removed without performance drop; actually, the performance is improved by a visible margin. This new dictionary-free MoCo framework is termed SimMoCo, where ``Sim'' stands for ``Simplified'' and indicates the removal of the dictionary. 

\textbf{Dual temperature and its rationale.} A key property of the SimMoCo is that it uses dual temperature for enabling independent control of intra-anchor and inter-anchor hardness-aware properties with $\tau_{\alpha}$ and $\tau_{\beta}$, respectively. As shown in Table~\ref{tab:DT_reverse}, adopting a sufficiently large $\tau_{\beta}$ reduces the inter-anchor hardness-aware property and boosts the performance. On the other hand, the vector component does not allow such a sufficiently large temperature. This creates a dilemma choice of setting an appropriate single temperature. The rationale discussed here is supported by the results in Table~\ref{tab:DT_reverse}. This rationale also somehow aligns with an observation in ~\cite{yeh2021decoupled} that removing the positive pair from the denominator in the InfoNCE loss for decoupling the influence of positive sample and negative ones on each other. ~\cite{yeh2021decoupled} justifies their decoupling from the perspective of learning efficiency, while our dual temperature highlights independent control of two hardness-aware properties.

\begin{table}[!htbp]
    \centering
    \resizebox{0.7\linewidth}{!}{ 
    \begin{tabular}{ccccc}
    \toprule
    Method & $\tau_{\alpha}$ & $\tau_{\beta}$ & Accuracy (\%)   \\
    \midrule
    ST & 0.1 & 0.1 &  49.52 \\
    ST & 1 & 1 & 32.09 \\
    DT & 0.1 & 1 & \textbf{ 54.11 }  \\
    DT (reverse) & 1 & 0.1 &   31.28 \\
    \bottomrule
    \end{tabular} }
    \caption{ Performance on CIFAR100 with different configurations of $\tau_{\alpha}$ and $\tau_{\beta}$. ST indicates \textit{single temperature} with $\tau_{\beta} = \tau_{\alpha}$. DT indicates \textit{dual temperature} with $\tau_{\beta} \neq \tau_{\alpha}$. DT (reverse) indicates a larger temperature is set for the vector component, \ie\ $\tau_{\alpha} > \tau_{\beta}$. 
    }
    \label{tab:DT_reverse}
\end{table}

\subsection{Momentum Removal}
\textbf{SimCo.} The momentum has been introduced in MoCo to increase the consistency of the dictionary~\cite{he2020momentum}. Our proposed SimMoCo already has no dictionary, thus it might make sense to further remove the momentum from the SimMoCo for simplicity. Another merit of removing such momentum is that it allows the gradient to backward through the side of keys, which is empirically found to boost performance. The simplified momentum-free variant of SimMoCo is straightforwardly termed SimCo. In the SimCo, $q$ and $k$ are from the same encoder and they are symmetric with different augmentations. Assuming that the MBS is N, $q$ and $k$ both have N elements, for which $q_i$ and $k_j$ are positive samples to each other when $i = j$; otherwise they are negative samples to each other. The loss in Eq~\ref{eq:infonce_nss_equiva} does not allow gradient on $k$. To enable the gradient update on both $q$ and $k$ , we propose an alternative implementation for dual temperature. 
An new variant of InfoNCE with dual temperature (DT) can be designed as:

\begin{equation}
\vspace{-0.05in}
\left\{
\begin{aligned}
& \mathcal{L}^{DT}_{q_i} = -\text{sg}(\frac{W^i_{\beta}}{W^i_{\alpha}}) \log \frac{\exp({q_i{\cdot}k_i /\tau_{\alpha}})}{\sum\limits_{j=1}^{N} \exp({q_i{\cdot}k_j/\tau_{\alpha}})} \\
W^i_{\beta} =1 -& \frac{\exp({q_i{\cdot}k_i /\tau_{\beta}})}{\sum\limits_{j=1}^{N} \exp({q_i{\cdot}k_j/\tau_{\beta}})}, W^i_{\alpha} = 1- \frac{\exp({q_i{\cdot}k_i /\tau_{\alpha}})}{\sum\limits_{j=1}^{N} \exp({q_i{\cdot}k_j/\tau_{\alpha}})},
\end{aligned}
\right.
\label{eq:alter_imple}
\end{equation}
where $\text{sg}(\frac{W^i_{\beta}}{W^i_{\alpha}})$ changes the temperature from $\tau_{\alpha}$ to $\tau_{\beta}$ for the scalar component, while keeping $\tau_{\alpha}$ in the vector component unchanged. Taking the symmetry into account, the final loss would be $(\mathcal{L}_{q_i} + \mathcal{L}_{k_i})/2$, where $\mathcal{L}_{k_i}$ has the same form as $\mathcal{L}_{q_i}$ but switches the position of $q$ and $k$. With gradient update on both $q$ and $k$, the loss in Eq~\ref{eq:alter_imple} resembles that in~\cite{chen2020simple} but uses half negative samples. More discussion on their relationship as well as the pseudo code for Eq~\ref{eq:alter_imple} are in the \textcolor{black}{supplementary}.

\section{Experimental Setup and Results.}
\subsection{Experimental Setup.}

\textbf{Training.} Following the settings on \textit{CIFAR experiment} in the official GitHub repository~\footnote{https://github.com/facebookresearch/moco}, we use SGD optimizer with momentum 0.9 and weight decay 5e-4, and the temperature is set to 0.1. We train each model for 200 epochs with the MBS of 256 on a single GPU. In the first 10 epochs, we use a linear warmup learning rate then decay learning rate following cosine decay schedule without restarts \cite{loshchilov2016sgdr}. The highest learning rate is set to 0.03. The momentum coefficient is set to 0.99. The projector of baseline MoCo v2 consists of two linear layers with 
a ReLU activation function between them, for which we keep the same setting. The augmentations adopted are random color jittering, random horizontal flip, and random grayscale conversion. We highlight that for a fair comparison, \textit{MoCo v2, SimMoCo, and SimCo are always trained under the same setup except for the specified changes, such as the intended dual temperature and removal of dictionary and momentum}. For the dual temperature, we need to set the temperature to different values. $\tau_{\alpha}$ needs to be set to an appropriate value due to the so-called uniformity-tolerance dilemma~\cite{Wang_2021_CVPR}. We follow common setups to set $\tau_{\alpha}$ to 0.1. For $\tau_{\beta}$, we set it to 1.0. Empirically, we find that $\tau_{\beta}$ has no significant influence on the performance as long as it is set to a sufficiently large value for mitigating the inter-anchor hardness-aware property.

\textbf{Evaluation.} As shown in the solo-learn~\cite{turrisi2021sololearn} frameworks, the performance gap between online and offline linear evaluation is not significant. For the convenience to avoid the need of retraining a linear classifier after the encoder pretraining, we directly report top-1 accuracy (\%) on the validation dataset with the online linear evaluation.

\subsection{Experimental Results}
\textbf{Temperature $\tau_{\alpha}$.} Temperature has been identified as an important hyperparameter for controlling the balance between uniformity and tolerance~\cite{Wang_2021_CVPR}.
With ResNet 18 on CIFAR100, the results with a wide range of $\tau_{\alpha}$ are reported in Table~\ref{tab:temp_vary}. For all the three frameworks, a very small or a very large $\tau_{\alpha}$ leads to inferior performance. Relatively, however, MoCo v2 is more sensitive to 
temperature variation, for which a detailed discussion is provided in the \textcolor{black}{supplementary}.

\begin{table}[!htbp]
    \centering
    \resizebox{0.7\linewidth}{!}{ 
    \begin{tabular}{ccccc}
    \toprule
    $\tau_\alpha$ & 0.05 & 0.1 & 0.5 & 1  \\
    \midrule
    MoCo V2 & 49.16 & 53.28 & 35.99 & 21.74 \\
    \midrule
    SimMoCo & 53.67 & 54.11 & 42.25 & 32.42 \\
    SimCo   & 56.95 & \textbf{58.35} & 48.98 & 39.49 \\
                    \bottomrule 
    \end{tabular} }
    \caption{Performance under different temperature settings of $\tau_\alpha$ on the scalar component. } 
    \label{tab:temp_vary}
\end{table}

\begin{table}[!htbp]
    \centering
    \resizebox{0.8\linewidth}{!}{ 
    \begin{tabular}{cccccc}
    \toprule
    Batch size & 64 & 128 & 256 & 512 & 1024  \\
    \midrule
    MoCo v2 & 52.58 & 54.40 & 53.28 & 51.47 & 48.90 \\
    \midrule
    SimMoCo & 54.02 & 54.93 & 54.11 & 52.45 & 49.70 \\
    SimCo & 58.04 & 58.29 & \textbf{58.35} & 57.08 &  55.34  \\
                                                    \bottomrule 
    \end{tabular} }
    \caption{Performance comparison with different mini-batch sizes on CIFAR-100.
    } 
    \label{tab:batchsize_res}
\end{table}

\textbf{Mini-batch size.} Here, we further investigate another important hyperparameter, mini-batch size (MBS) with the linear-scaling rule~\cite{goyal2017accurate} adopted to change the learning rate proportional to the MBS. As shown in Table~\ref{tab:batchsize_res}, we can observe that the proposed SimMoCo and SimCo achieve superior performance over a wide range of MBS. Notably, a smaller MBS leads to inferior performance for all the three frameworks, while a larger MBS does not always lead to a better performance, which can be attributed to training difficulty with a large MBS~\cite{you2017large}.

\textbf{Longer training.} We experiment with longer training and the results are shown in Table~\ref{tab:Longer_training}. The superiority of our simplifed frameworks over the MoCo v2 can also be observed for longer epochs. 
\begin{table}[!htbp]
\centering
\resizebox{0.6\linewidth}{!}{ 
\begin{tabular}{cccc}
\toprule
Epoch & 200 & 400 & 800 \\
\midrule
MoCo v2 & 53.28  & 59.7 & 63.74  \\ 
\hline
\textbf{SimMoCo} & 54.11   & 60.11  & 63.82  \\ 
\textbf{SimCo} & 58.35   & 62.36  & 65.68  \\ 
\bottomrule
\end{tabular}
} 
\caption{Performance comparison for longer training.} 
\label{tab:Longer_training}
\end{table}

\textbf{Various architectures.} On CIFAR100, we further compare the three frameworks with different architectures, including  ResNet18, ResNet50, ViT tiny \cite{dosovitskiy2021an}, Swin tiny \cite{liu2021swin}.  The results in Table~\ref{tab:resnet50_res} suggest that SimMoCo consistently outperforms MoCo v2. 
SimCo consistently outperforms MoCo v2 as well as our SimMoCo by a large margin. 

\begin{table}[!htbp]
    \centering
    \resizebox{1.0\linewidth}{!}{ 
    \begin{tabular}{ccccc}
    \toprule
    Architecture & ResNet-50 & ViT tiny  & Swin tiny & ResNet-18 \\
    \midrule
    MoCo v2 & 53.44 & 16.78  &32  & 53.28  \\
        \midrule
    SimMoCo &54.64   & 21.35 & 33.07  & 54.11 \\
                    SimCo & \textbf{58.48}  & \textbf{28.81}  & \textbf{42.64} & \textbf{ 58.35 } \\
    \bottomrule 
    \end{tabular} }
    \caption{Performance comparison on CIFAR-100 with different architectures. 
        } 
    \label{tab:resnet50_res}
\end{table}

\textbf{Various dataset.} With ResNet18, we evaluate on multiple datasets, including CIFAR-10, CIFAR-100, SVHN, STL10 and ImageNet-100. The results in Table \ref{tab:dataset_varying} show that our simplified models consistently outperform the baseline MoCo v2, except that the performance of SimMoCo is slightly worse than that of MoCo v2. Despite the simplicity, SimCo generally performs the best among the three frameworks on all investigated datasets except for SVHN dataset, for which SimMoCo performs the best.

\begin{table}[!htbp]
    \centering
    \resizebox{0.99\linewidth}{!}{ 
    \begin{tabular}{cccccc}
    \toprule
    Dataset & CIFAR10 & CIFAR100 & SVHN & STL10 & ImageNet100  \\
    \midrule
    MoCo v2 & 82.35 &  53.28 &47.25  & 81.25  &  57.52 \\ 
        \midrule
        SimMoCo & 82.36 & 54.11 & 53.67 & 80.56   &  58.2  \\ 
        SimCo & \textbf{85.61} & \textbf{58.35} &52.37  & \textbf{83.19} & \textbf{61.28} \\ 
    \midrule
    MoCo v2+ & 85.3 &  57.19 & 44.51 & 82.93 & 60.52  \\ 
    SimMoCo+ & 85.61 & 58.15 & 44.74 & 82.61 & 61.12  \\ 
    \bottomrule
    \end{tabular} }
    \caption{Results w/ or w/o symmetric loss on various datasets. Note that by default SimCo adopts a symmetric loss.}
    \label{tab:dataset_varying}
\end{table}

\textbf{Symmetric MoCo and SimMoCo.}~\cite{chen2021empirical} has shown a symmetric loss leads to a performance boost for the frameworks with MoEn. Following~\cite{chen2021empirical}, we term them MoCo v2+ and SimMoCo v2+ when the symmetric loss is adopted. The results in Table~\ref{tab:dataset_varying} show that the performance is boosted by a large margin. It is worth highlighting that unlike SimCo with a default symmetric loss, the symmetric loss in the MoEn-based MoCo v2+ and SimMoCo+ doubles the computation resources. Nonetheless, SimCo still outperforms them by a visible margin.

\section{A Unified Perspective on SSL and Beyond}
\begin{figure}[!htbp]
  \centering
 \includegraphics[width=1\linewidth]{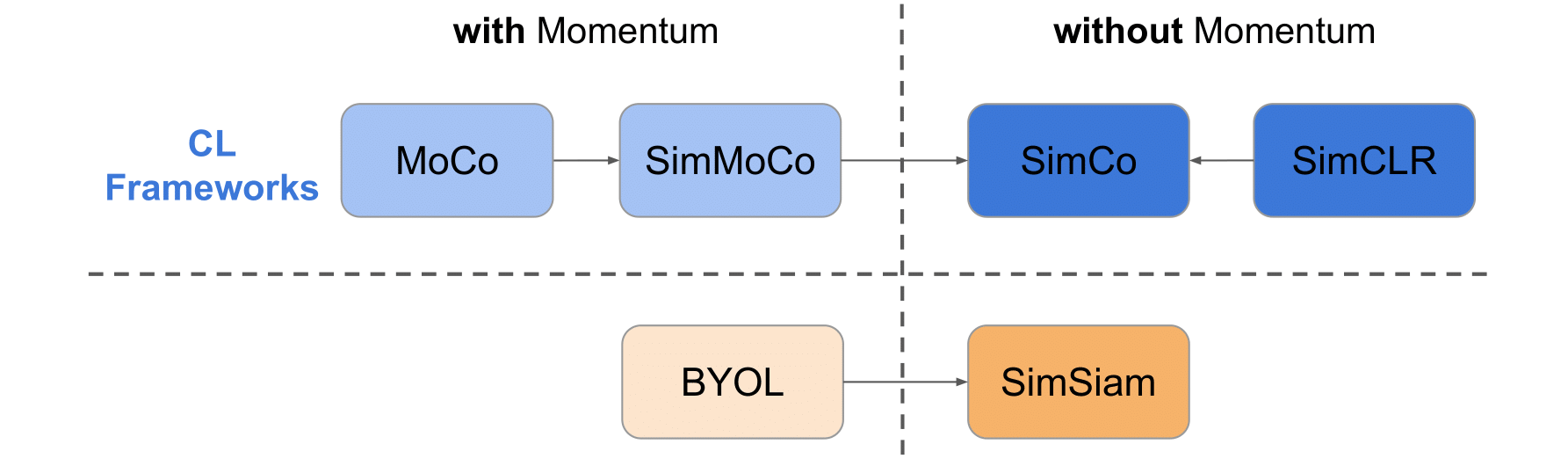}
  \caption{
  Comparison between CL and non-CL frameworks.
  }
  \label{fig:framework_bridge}
\end{figure}

Currently, the CL frameworks can be roughly divided into two categories: (a) dictionary-free CL represented by SimCLR~\cite{chen2020simple} and (b) dictionary-based CL represented by MoCo family. As shown in Figure~\ref{fig:framework_bridge}, our proposed SimMoCo simplifies MoCo via dictionary removal and is further simplified into SimCo via momentum removal. SimCo and SimCLR are both dictionary-free and momentum-free, and SimCLR can be roughly perceived as a special case of SimCo with the dual temperature set to the same value. In the following, we discuss how our investigation further brings a unified perspective on CL and non-CL frameworks. 

\begin{table}[!htbp]
    \centering
    \resizebox{0.9\linewidth}{!}{ 
    \begin{tabular}{cccccc}
    \toprule
    Type & Methods &  Inter-anchor HA  & Top-1 (\%)  \\
    \midrule
    \multirow{5}{*}{CL}
        & SimMoCo &  \cmark & 50.22 \\
    & SimMoCo & \xmark  & 54.11 \\
    & SimCo &  \cmark & 53.82 \\
    & SimCo & \xmark  & 58.35 \\
   \midrule
    \multirow{4}{*}{Non-CL}
        & BYOL & \cmark & 46.54 \\
    & BYOL  & \xmark & 50.65 \\
    
                    & SimSiam &  \cmark & 39.18 \\ 
    & SimSiam &  \xmark & 51.78 \\
    \bottomrule 
    \end{tabular} }
    \caption{Influence of Inter-anchor hardness-awareness (HA) in CL and non-CL frameworks. All methods are based on ResNet-18, trained on  CIFAR-100 with 200 epochs under the same setup.}         \label{tab:othermethod}
\end{table}

\subsection{Bridging the Gap Between CL and Non-CL} 
The SSL frameworks can be divided into CL and non-CL based on whether negative samples are used. For simplicity, we first discuss non-CL frameworks that use a simple loss:
\begin{equation}
    \mathcal{L}^{ncl}_{q_i} =  h(q_i) \cdot \text{sg}[-k_i],
    \label{eq:loss_noncl}
\end{equation}
where $h$ is a prediction head. The above loss is adopted in BYOL~\cite{grill2020bootstrap} and SimSiam~\cite{chen2021exploring}, which are arguably the two most popular non-CL frameworks. In practice, they use a symmetric loss and here for notation simplicity, we only take the non-symmetric loss into account. Comparing the loss in Eq~\ref{eq:infonce_nss_equiva} and that in Eq~\ref{eq:loss_noncl}, we note a difference: the vanilla contrastive InfoNCE puts different penalty weights on anchors, while the non-CL frameworks~\cite{chen2021exploring,grill2020bootstrap} adopt a loss that treats all anchors equally. 
 
With the gradient on $k$ disabled and $\tau_{\beta}$ set to sufficiently large, Eq~\ref{eq:alter_imple} can be reformulated in the form of Eq~\ref{eq:infonce_nss_equiva} as:
\begin{equation}
    \mathcal{L}^{cl}_{q_i} =  q_i \cdot \text{sg}[(-k_i + \sum_{j=1}^{K} \hat{\bm{p}}^i_j k_j)].
    \label{eq:loss_cl}
\end{equation}
where $\frac{1}{\tau_{\alpha}}$ is omitted for simple discussion as aforementioned. It is interesting to note that the above loss resembles Eq~\ref{eq:simple} for treating all anchors equally and resembles Eq~\ref{eq:infonce} for keeping the intra-anchor hardness-awareness. Recently,~\cite{zhang2022how} has shown that the negative samples in CL frameworks and predictor $h$ in non-CL frameworks, SimSiam~\cite{chen2021exploring} for instance, achieve equivalent roles of de-centering and de-correlation for avoiding collapse. Some recent non-CL frameworks~\cite{zbontar2021barlow,bardes2021vicreg} replace predictor with explicit de-correlation and regularization. 
In other words, their finding mainly bridges the gap between CL and non-CL frameworks from the perspective of intra-anchor hardness-awareness. Our work fills the gap by pointing out that the loss in non-CL frameworks~\cite{chen2021exploring,grill2020bootstrap,ermolov2021whitening,zbontar2021barlow,bardes2021vicreg} treats anchors equally, while vanilla contrastive InfoNCE in CL penalizes each anchor based on their hardness. Overall, through alleviating this imbalance (see the weight on $k_i$ in Eq~\ref{eq:loss_cl}), our work further bridges the gap between CL and non-CL frameworks to have a unified understanding of SSL. 

\textbf{Inter-anchor hardness-awareness in non-CL.} It is interesting whether inter-anchor hardness-awareness also affects non-CL frameworks. As aforementioned, non-CL frameworks treat anchors equally and thus we modify the loss in Eq~\ref{eq:loss_noncl} via multiplying it by $\text{sg}[\sum_{j=1}^{K}\bm{p}^i_j]$ for introducing inter-anchor hardness-awareness into non-CL. The results in Table~\ref{tab:othermethod} shows that such hardness-awareness also hurts the performance of both CL and non-CL frameworks.

\subsection{Discussion}

\textbf{Inter-anchor hardness-awareness in SSL \textit{vs.} SL.} With the softmax function, cross-entropy (CE) loss in supervised learning (SL) also has the inter-anchor hardness-aware property. Our investigation suggests that unlike InfoNCE in SSL, such property is critical for competitive performance in SL. We find that this can be partly attributed to the explanation that this default anchor-wise weight is less reliable to indicate the hardness than that in SL. A more detailed discussion is provided in the \textcolor{black}{supplementary}.

\section{Conclusion}

In this work, we revisit MoCo family by analyzing its key component, namely momentum-based dictionary. Our extensive analysis reveals that such a large dictionary is required mainly due to an inter-anchor hardness-awareness property of the commonly used InfoNCE in CL. We propose to control two hardness-aware properties independently with dual temperature, which facilitates simplifying MoCo v2 through removing the dictionary as well as momentum. Extensive experiments have confirmed that our simplified frameworks, SimMoCo and SimCo, achieve competitive performance against their baseline MoCo V2. This work also bridges the gap between CL and Non-CL frameworks to form a unified understanding of SSL.

\section*{Acknowledgement}
This work was partly supported by Institute for Information \& communications Technology Planning \& Evaluation (IITP) grant funded by the Korea government (MSIT) under grant No.2019-0-01396 (Development of framework for analyzing, detecting, mitigating of bias in AI model and training data), No.2021-0-01381 (Development of Causal AI through Video Understanding and Reinforcement Learning, and Its Applications to Real Environments) and No.2021-0-02068 (Artificial Intelligence Innovation Hub).

{
        \small
    \bibliographystyle{ieee_fullname}
    \bibliography{macros,bib_mixed}
}

\newpage

\appendix

\setcounter{page}{1}

\twocolumn[
\centering
\Large
\textbf{[CVPR2022]} \\
\vspace{0.5em}Supplementary Material \\
\vspace{1.0em}
] 
\appendix

\section{Setup of Figure 2 in the main manuscript}
\begin{figure}[!htbp]
  \centering
 \includegraphics[width=1\linewidth]{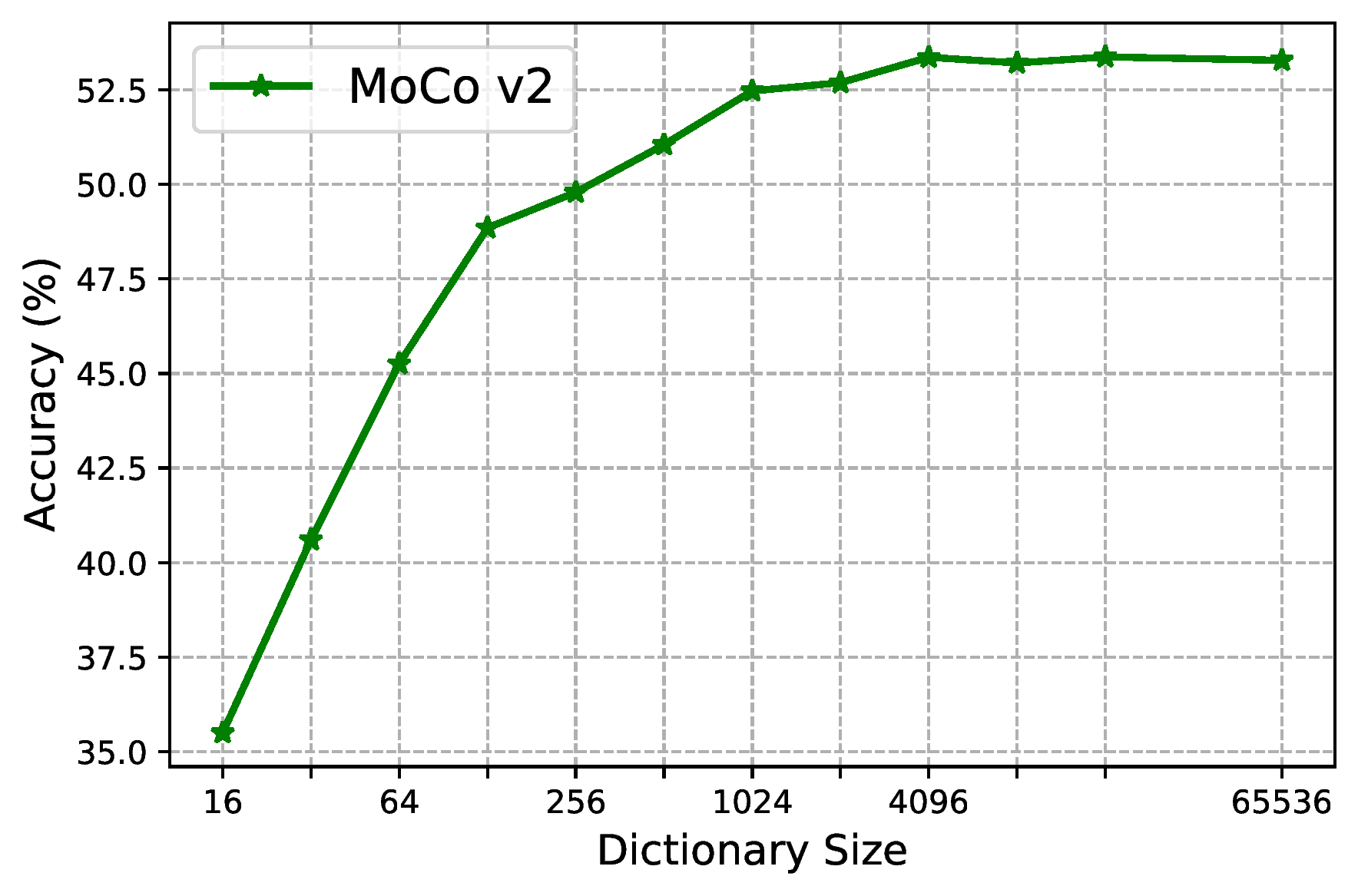}
  \caption{Influence of dictionary size in MoCo v2.}
  \label{fig:interintra_both}
\end{figure}

We train the model on CIFAR100 with MoCo v2 for 200 epochs on a single GPU. we use SGD optimizer with momentum 0.9 and weight decay 5e-4, and the temperature is set to 0.1. We use a linear warmup learning rate then decay learning rate following cosine decay schedule without restarts. Here, we adopt two independent dictionaries, $\bm{D}_{vector}$ and $\bm{D}_{scalar}$ to store negative sample keys for vector and scalar components, respectively. We fix one of them to have the dictionary size of 65536, while changing the dictionary size of the other one. We also report the results of a single dictionary with various dictionary size in Fig.~\ref{fig:interintra_both}. As expected, the performance decreases significantly when the dictionary size is small.

\section{The scalar component is less sensitive to the quality of the keys}
We also report the results for sampling a certain number ($K_{scalar}$) of keys from $\bm{D}_{scalar}$ while using the full $\bm{D}_{vector}$. We set $K_{scalar}$ to 4096, since our results in Figure 2 of the main manuscript show that the scalar component requires a sufficiently large dictionary for competitive performance. The results in Table~\ref{tab:comp_quality_2} show that there is only a small performance gap among the three sampling strategies. Notably, the model still converges well with a reasonable performance even when the earliest keys are sampled, while the model does not converge for $K_{vector}$ in the same setup. The results show that the scalar component is less sensitive to the key quality. 

\begin{table}[!htbp]
    \centering
    \begin{tabular}{ccccc}
    \toprule
    Sampling strategies & Earliest & Random & Newest \\
    \midrule
        Top-1 Accuracy(\%)  & 52.13  & 52.75  & 53.32  \\
                                                \bottomrule
    \end{tabular}
    \caption{Comparison of various sampling strategies on CIFAR100.}
        \label{tab:comp_quality_2}
\end{table}

\section{The pseudo code for the relationship of dual temperature}
The core difference between the InfoNCE with dual temperature in Eq 9 of the main manuscript and that in~\cite{chen2020simple} lies in whether dual temperature is applied. Moreover, the loss in~\cite{chen2020simple}  uses negative samples from both encoders, while InfoNCE with dual temperature uses only half negative sample. For example, when $q_i$ is the anchor, it only uses negative samples from the encoder $k$ side, which simplifies the code implementation. The pseudo code is shown in Algorithm~\ref{alg:mirror_asymmetric}. Adopting negative samples from both sides is confirmed to yield equivalent performance.

\begin{algorithm}[ht]
\caption{Pytorch-like Pseudocode: Dual Temperature Loss}
\label{alg:mirror_asymmetric}
\definecolor{codeblue}{rgb}{0.25,0.5,0.5}
\definecolor{codekw}{rgb}{0.85, 0.18, 0.50}
\lstset{
  backgroundcolor=\color{white},
  basicstyle=\fontsize{7.5pt}{7.5pt}\ttfamily\selectfont,
  columns=fullflexible,
  breaklines=true,
  captionpos=b,
  commentstyle=\fontsize{7.5pt}{7.5pt}\color{codeblue},
  keywordstyle=\fontsize{7.5pt}{7.5pt}\color{codekw},
}
\centering
\begin{lstlisting}[language=python]

def simco_loss(query, key, intra_temperature, inter_temperature):
    """
    N: batch size
    D: the dimension of representation vector
    
    Args:
        query (torch.Tensor): NxD Tensor containing projected features from view 1.
        key (torch.Tensor): NxD Tensor containing projected features from view 2.
        intra_temperature (float): temperature factor for the intra component.
        inter_temperature (float): temperature factor for the inter component.

    Returns:
        torch.Tensor: SimCo loss.
    """
    # normalize query and key
    query = F.normalize(query, dim=-1) 
    key = F.normalize(key, dim=-1)
    
    # calculate logits
    logits = query @ key.T
    
    # intra awareness 
    logits_intra = logits / intra_temperature
    prob_intra = F.softmax(logits_intra, dim=1)
    
    # inter awareness
    logits_inter = logits / inter_temperature
    prob_inter = F.softmax(logits_inter, dim=1)
    
    # inter awareness changing factor
    mask = torch.ones(prob_inter.size()).fill_diagonal_(0)
    weight_alpha = (prob_intra * mask).sum(-1)
    weight_beta = (prob_inter * mask).sum(-1)
    
    inter_intra = weight_beta / weight_alpha
    
    # loss calculation
    log_softmax = F.log_softmax(logits, dim=-1)
    log_softmax_diag = log_softmax.diag()
    
    loss = -inter_intra.detach() * log_softmax_diag 
    return loss.mean()

\end{lstlisting}
\end{algorithm}

\section{MoCo v2 is more sensitive to temperature variation}
Note that MoCo v2 by default adopts a single temperature, \ie\ $\tau_{\beta} = \tau_{\alpha}$. When the temperature is very small, the inter-anchor hardness-aware sensitivity gets higher, leading to lower performance, while our SimMoCo and SimCo have no such concerns because $\tau_{\beta}$ is large. When the temperature is very large, the dependence of MoCo v2 on the old keys gets higher, \ie\ lower PN consistency. The PN consistency for our SimMoCo and SimCo is always optimal because the negative keys are generated by the same encoder as the positive keys. Thus, our SimMoCo and SimCo have no such consistency concerns as MoCo v2. Overall, we observe that our proposed SimMoCo and SimCo consistently outperform the baseline MoCo v2.

\section{InfoNCE in SSL \textit{vs.} CE in SL.} 

The CE loss in supervised learning (SL) is shown as 
\begin{equation}
    \mathcal{L}_{CE} = -\log \frac{\exp({\bm{o}_{gt}/\tau})}{\sum_{c=1}^{C} \exp({\bm{o}_c/\tau})},
    \label{eq:ce}
\end{equation}
where $\bm{o}$ indicates the network output which is a logit vector of length $C$ (total number of classes) and ${gt}$ indicates the index for the ground-truth (GT) class. Note that the sum is over the GT class and $(C-1)$ non-GT classes. With one hot vector defined as $\bm{y}$, there exists the following equivalence: $\bm{o}_{gt} = \bm{o} \cdot \bm{y}_{gt}$ and $\bm{o}_{c} = \bm{o} \cdot \bm{y}_{c}$. 

Based on the above equivalence, compared with Eq 1 in the main manuscript, we show that CE loss is a special case of InfoNCE by perceiving the GT one-hot vector as the positive key and other non-GT one-hot vectors as negative keys. With such a high resemblance between the two losses, however, unlike InfoNCE in SSL, this inter-anchor hardness-aware property is widely known to be important for competitive performance. In other words, alleviating the inter-anchor hardness-ware property does not help CE loss to improve the performance. 

\begin{figure}[!htbp]
  \centering
 \includegraphics[width=1\linewidth]{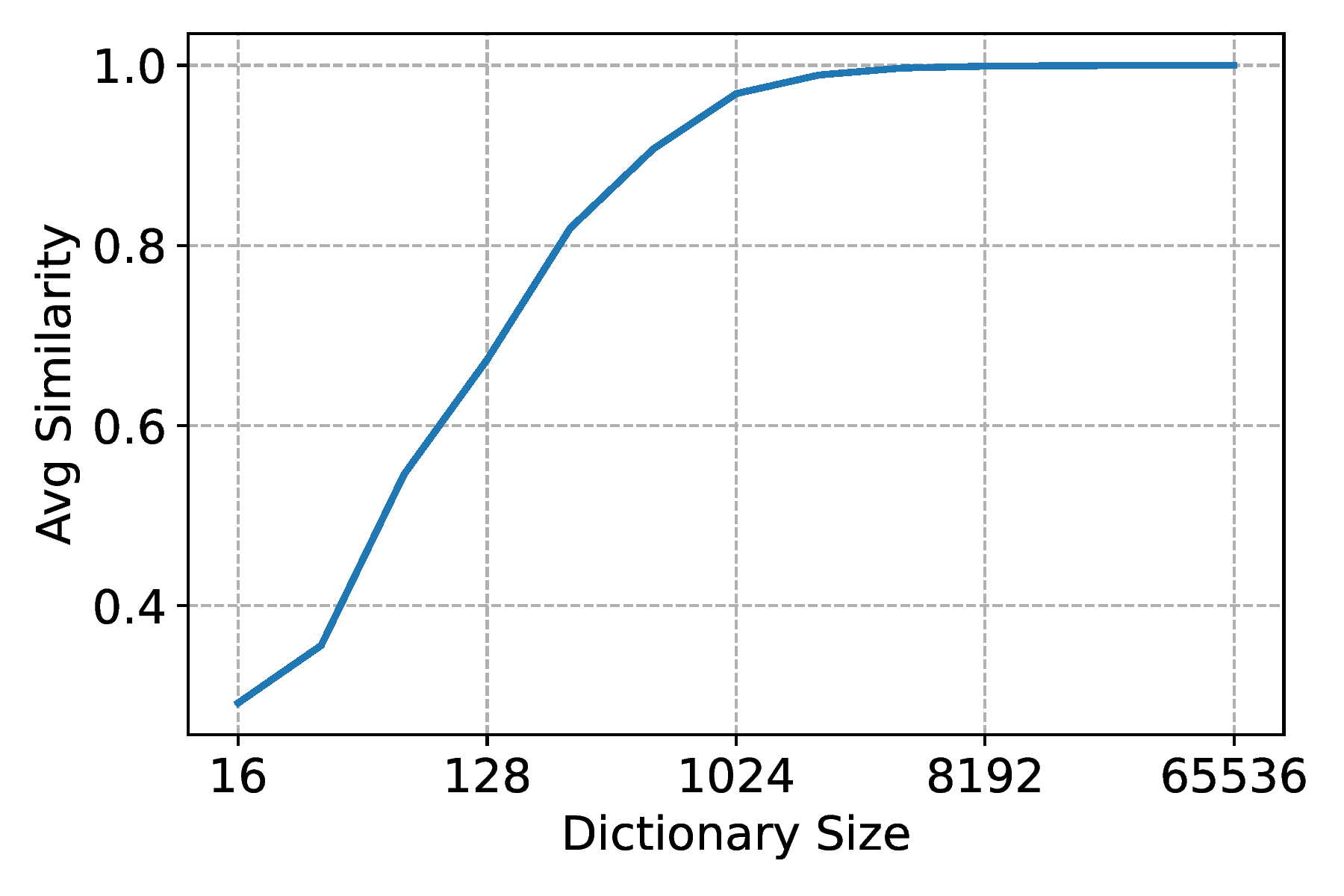}
  \caption{Cosine similarity between different $r^i_+$ through changing the positive and negative keys randomly. Low similarity indicates that the inter-anchor hardness-aware weight is not reliable because a reliable prior should not deviate too much through changing the positive and negative keys.}
  \label{fig:abg_similarity}
\end{figure}

Here, we attempt to provide an intuitive explanation. Imagine that we do not have prior knowledge on the hardness of anchor sample, straightforwardly, the loss should be designed to treat every anchor sample equally. Given such prior knowledge, it is intuitive that the loss should put more weight on the hard anchor samples, such as CE does. Regarding this prior, the main difference between InfoNCE and CE is that the prior knowledge in CE is very reliable because the keys (both GT and non-GT) are fixed yet correct. However, this prior is less reliable in the InfoNCE loss because the keys are random. For example, the positive key with the same image of another random augmentation, and the negative keys are encoded from the random images. By changing the positive and negative keys randomly, we get two sets of $r^i_+$ (see Eq 8 in the main manuscript) and calculate their similarity. The results in Figure~\ref{fig:abg_similarity} show that the similarity is low when the dictionary size is small, indicating this inter-anchor weight is not reliable. Intuitively, if this prior is unreliable, this inter-anchor hardness-aware property is misleading and thus it might be better to decrease this hardness-aware property, \ie\ treating every anchor sample equally as in our investigation.  

\begin{table}[!htbp]
\centering
\resizebox{\linewidth}{!}{ \begin{tabular}{c|ccc|ccc}
\toprule
 \multirow{2}{*}{Method}   & \multicolumn{3}{c|}{Symmetric} & \multicolumn{3}{c}{Asymmetric} \\
           & 0.4       & 0.6      & 0.8  & 0.4       & 0.6      & 0.8   \\
\hline
 CE       &   57.59    &   39.36   &  20.39 & 57.89  &   38.62       &  19.29  \\
 CE (DT)   &  \textbf{63.95}    &  \textbf{56.21}    &\textbf{22.51} & \textbf{63.07}          &  \textbf{59.53}       & \textbf{21.8}    \\
\bottomrule 
\end{tabular}
}
\caption{
Test accuracy (\%) of standard CE and CE (DT) on CIFAR10 with symmetric label noise ($\eta\in\{0.4,0.6,0.8\}$) and asymmetric label noise ($\eta\in\{0.4,0.6,0.8\}$). 
} \label{tab:ablations}
\end{table}

With the above interpretation, the inter-anchor hardness-aware weight might also be detrimental to CE loss if the prior gets less reliable. A straightforward way to make the prior less reliable is to corrupt the data with noisy labels. We follow the setup in prior works~\cite{ma2020normalized} that study noisy labels. Specifically, the noise can be corrupted in a symmetric or asymmetric manner. The results with different noise ratios are shown in Table~\ref{tab:ablations}. We observe that CE with dual temperature to remove the inter-anchor hardness-aware property outperforms the standard CE loss by a visible margin. Note that this experiment is conducted to prove our interpretation instead of pushing the SOTA performance in the setup of noisy labels.

\end{document}